# Generalization Enhancement Strategies to Enable Cross-year Cropland Mapping with Convolutional Neural Networks Trained Using Historical Samples

**Authors:**

Sam Khallaghi[1,2,*], Rahebe Abedi[1], Hanan Abou Ali[3], Hamed Alemohammad[1,2], Mary Dziedzorm Asipunu[4], Ismail Alatise[1], Nguyen Ha[1], Boka Luo[1], Cat Mai[1], Lei Song[1,5], Amos Wussah[4], Sitian Xiong[1], Yao-Ting Yao[1,2], Qi Zhang[1], Lyndon D. Estes[1]

**Affiliation:**

[1]Graduate School of Geography, Clark University, Worcester, MA USA

[2]Clark Center for Geospatial Analytics, Clark University, Worcester, MA USA

[3]Department of Geography and Spatial Sciences, University of Delaware, Newark, DE

[4]Farmerline Ltd., Kumasi, Ghana

[5]Department of Geography, University of California Santa Barbara, Santa Barbara, CA USA

*Corresponding author.

## Abstract

The accuracy of mapping agricultural fields across large areas is steadily improving with high-resolution satellite imagery and deep learning (DL) models, even in regions where fields are small and geometrically irregular. However, developing effective DL models often requires large, expensive label datasets, typically available only for specific years or locations. This limits the ability to create annual maps essential for agricultural monitoring, as domain shifts occur between years and regions due to changes in farming practices and environmental conditions. The challenge is to design a model flexible enough to account for these shifts without needing yearly labels.

While domain adaptation techniques or semi-supervised training are common solutions, we explored enhancing the model's generalization power. Our results indicate that a holistic approach is essential, combining methods to improve generalization. Specifically, using an area-based loss function, such as Tversky-focal loss (TFL), significantly improved predictions across multiple years. The use of different augmentation techniques helped to encode different types of invariance, particularly photometric augmentations encoded invariance to brightness changes, though they increased false positives. The combination of photometric augmentation, TFL loss, and MC-dropout produced the best results, although dropout alone led to more false negatives in subsequent year predictions. Additionally, the choice of input normalization had a significant impact, with the best results obtained when statistics were calculated either locally or across the entire dataset over all bands (lab and gab). We developed a workflow that enabled a U-Net model to generate effective multi-year crop maps over large areas. Our code, available at: https://github.com/agroimpacts/cnn-generalization-enhancement, will be regularly updated with improvements.

## 1. Introduction

The ability to produce precise, field-level maps of agricultural lands is crucial for understanding and managing the complex interplay between food security, land use, and environmental sustainability (Ajadi et al. 2021). These maps are key to understanding farmland configurations, including the spatial distribution, quantity, morphology, and dimensions of fields, which serve as pivotal indicators in domains such as land management (Bhosle and Musande, 2019), ecosystem monitoring (Burkhard et al. 2012, Akbari et al. 2020), food security (Zhang et al. 2018, Karthikeyan et al. 2020), and precision agriculture (Mazzia et al. 2020, Khan et al. 2021).

Such a growing need for extensive, routine, and automated mapping of crop fields is driven by the rapid evolution of global agricultural systems and food security challenges in the face of climate change. This need is especially pressing in smallholder-dominated regions of Asia and Africa, where landscapes are typically characterized by small (less than 1-2 hectares), geometrically irregular, and dynamically changing fields, often featuring heterogeneous management practices and frequent presence of trees within the fields (Lambert et al. 2018, Jin et al. 2019, Estes et al. 2022).

Available global or continental-scale cropland maps are typically coarse in resolution and produced infrequently with poor or unverified accuracy (e.g. Global data set of monthly irrigated and rainfed crop areas around the year 2000 (MIRCA2000), the Global Rain-fed, Irrigated, and Paddy Croplands (GRIPC) (Salmon et al. 2015) while frequent large-scale products are usually limited to developed countries (e.g. U.S. Cropland Data Layer (CDL) or Canadian Annual Crop Inventory) (Waldner et al. 2015). The Global Food Security-support Analysis Data (GFSAD) project products like the Landsat-derived Global Rainfed and Irrigated-Area Product (LGRIP30) and the Landsat-derived global cropland extent product (LGCEP30) partially addresses some limitations in cropland data, such as imprecise spatial locations, and uncertainties in differentiating irrigated and rainfed areas. Another exception to this trend is the newly available LULC product of Dynamic World (Brown et al. 2022) at 10 meters resolution and near real-time global coverage but the resolution of these products is still too coarse to map many cultivated areas dominated by smallholder agricultural systems.

In the context of small-holder agricultural regimes, pivotal advances in cropland mask generation have arisen from the use of high spatial resolution imagery (<5 m), particularly when paired with improved revisit frequency (Wang et al. 2022). Although the use of unsupervised methods is increasing, prompted in part by the promising results from emerging foundation models that eliminate the necessity of annotated datasets in the training process, such as Prithvi (Jakubik et al. 2023) and Presto (Tseng et al. 2023), supervised learning techniques still continue to dominate the majority of research in this area. In addition to precisely identify and map the locations of cultivated pixels, there is an increasing emphasis on models that can accurately delineate the geometric contours of agricultural fields, to capture their shapes and sizes, which provides crucial information on agricultural systems while enhancing the overall accuracy of the mapping process (Xie et al. 2019, Sharifi et al. 2022, Tetteh et al. 2023).

Typically, any DL model optimized for object detection or semantic segmentation can be adopted and modified for the aim of crop extent mapping. Examples include DeepLabv3+ (Du et al. 2019), specifically designed models such as ResUNet-a (Diakogiannis et al. 2020), HRRS-U-Net (Xie et al. 2023), and ensemble models like CCTNet (Wang et al. 2022). A popular approach to improve segmentation masks is multi-task training, where the delineation of field boundaries serves as an auxiliary task to aid the primary segmentation goal, commonly designed as a multi-branch network with task-specific branches and strategic fusion methodologies to improve the crop mask (Wang et al. 2020, Shunying et al. 2023, Long et al. 2022, Xu et al. 2023, Luo et al. 2023). Complementing these methods, implementing boundary-aware loss functions has proven to be an effective strategy in refining the accuracy of field boundary detection (Li et al. 2023). Alternative strategies focused on field boundary delineation commonly follow a two-step process involving semantic segmentation followed by instance segmentation post-processing, as exemplified by the Ultrametric Contour Map (UCM) approach (Persello et al. 2019). End-to-end instance segmentation strategies, such as E2EVAP (Pan et al. 2023), and region-based CNNs such as mask-R-CNN (Mei et al. 2022) are also used for this purpose.

Despite the promising capabilities of deep learning models in classifying croplands and delineating fields, consistently achieving high performance across extensive spatial and temporal scales remains a significant challenge. Notably, these models often necessitate extensive training datasets with representative samples to ensure accuracy (Schmitt et al., 2019, Fu et al. 2023), a requirement that is difficult to meet in smallholder systems where ground or census data are sparse or unavailable due to collection costs and other resource constraints (Lesiv et al., 2019, Wang et al. 2022). Moreover, the temporal specificity of samples needed to classify seasonal crops means that, even if they are available, they are typically relevant only for the current season, thereby increasing the cost and complexity of annual crop mapping (Zhang et al. 2018, Kou et al. 2023). To address these challenges, there is a growing emphasis on reusing historical samples, either from different years or geographic locations, by employing knowledge transfer strategies to mitigate the effects of domain shift (Hao et al. 2020, Van den Broeck et al. 2022, Antonijević et al. 2023, Pandžić et al. 2024).

Domain shift in the context of agricultural mapping can be roughly categorized into two types: temporal and geographical. Temporal domain shift usually involves the changes in the marginal distribution of brightness values for each band in the input samples from the source year (e.g. training year) to the target years (i.e. covariate shift) (Ma et al. 2024). Such spectral discrepancies that arise over time within the same geographic location can be attributed to a variety of factors, including year-to-year variations in agricultural practices, such as crop rotation, intercropping, and changes in crop species. Environmental changes also play a role, such as fluctuations in soil moisture levels, weather patterns, and even subtle shifts in atmospheric conditions, by altering the spectral signatures of the land. These shifts lead to compromised prediction quality from one year to the next, particularly when gathering images from the same season in different years is not an option, e.g. due to cloud occlusion (Persello et al. 2019). Geographical domain shift, on the other hand, occurs when models developed and trained in one geographic region are applied to a different region, which, besides covariate shift, can also lead to a shift resulting from the label space differing between the source and target domains (Ma et al. 2024). This type of shift is driven by changes in the dominant landscape features and their configuration and regional differences in agricultural practices, which are often influenced by local agronomic, climatic, and cultural factors. Additionally, inherent environmental variations between regions, such as soil types, climate conditions, and regional atmospheric and illumination characteristics, lead to distinct spectral signatures.

In this manuscript we focus on addressing the challenges of the temporal domain shift to maintaining model accuracy, and on improving the ability to reuse historical samples within the same geographic area. Our goal is to improve the ability to develop reliable, annual, high-resolution maps of cropland characteristics at regional to national scales. Research on temporal generalization and the reuse of historical data in cropland mapping has been relatively limited. The majority of existing studies adopt a pixel-based approach, relying on time-series data and typically focusing on crop-type mapping, often using conventional machine learning algorithms (Jiang et al. 2022, Kou et al. 2023). In cross-year crop mapping, common practices involve the use of measures of spectral similarity between source and target years, or incorporating domain knowledge to provide contextual understanding. Both approaches are based on the premise that, despite year-to-year variations, certain spectral characteristics remain consistent and can be used to identify similar crop pixels across different years. Techniques such as Spectral Angle Distance (SAD) (Kruse et al. 1993) and Euclidean Distance (ED) metrics are examples of this approach, as used by Huang et al. (2020) to enhance global land cover mapping accuracy. Waldner et al. (2015) leverage temporal features derived from domain knowledge of the crop growth cycle profiles and their spectral characteristics, in order to improve the accuracy of crop type classification and mapping across different years. Liu et al. (2022) used local similarity between time-series spectral feature vectors from historical and target year samples as a basis for creating transferable training datasets. Ge et al. (2021) used a mixture of temporal and geographical domain adaptation by applying a phenological matching technique to adapt a U-Net, initially trained on rice and corn fields in the south-central US, for mapping these crops in the midwestern US and Northeast China. While these temporal generalization methods have shown promise, they typically require time-series data and extensive domain knowledge. To create a binary cropland mask, however, mono-temporal imagery is often sufficient, which eliminates the need for complex time-series analysis while minimizing the computational burden. These limitations have prompted us to seek alternative strategies to enhance model generalization, using a similar rationale to that of Wang et al. (2022), who used a task-specific model called FracTAL-ResUNet (Diakogiannis et al. 2021) to improve generalization while relying on weak supervision using imperfect labels to overcome the barrier of annotation scarcity.

To improve generalization, different aspects of a pipeline such as data pre-processing, model architecture, loss function, optimization, and regularization techniques play an important role. With the lack of time-series input for phenological mapping and the desire to develop a strategy that can be plugged into any supervised segmentation pipelines, we mainly emphasize input pre-processing and regularization techniques.

Input normalization is a standard pre-processing procedure in DL models, intended to standardize input features' magnitude to enhance training. Common normalization methods include z-value standardization and min-max normalization, typically applied per band across the entire training dataset. Since the input only interacts with the weights of the initial network layer, the impact of input normalization on model generalization remains not fully explored (Huang et al. 2023), and many research papers fail to adequately document the procedure, while only a few studies have directly addressed the effects of normalization on model output and generalizability. Pelletier et al. (2019) investigated various normalization methods for time series data in remote sensing. They observed that the z-value, calculated per time stamp or entire time series, could distort temporal profiles and obscure vegetation differences. To counter these limitations, they proposed a global feature min/max normalization using 2% and 98% percentiles, which better preserved temporal profile shapes. There are also attempts to remove the local trends. For instance, Nguyen et al. (2020) used patches of Landsat 8 time-series normalized by mean-centering each band based on the pixel values for each local tile to map paddy fields at the pixel level.

Image augmentation is another standard procedure to increase the training dataset by creating new samples through transformation of the original samples, which can also act as a regulator by increasing the variability of the dataset. There are many transformations introduced in the literature, ranging from weak geometric transformations, such as random cropping and re-scaling, and different types of flip and rotation that preserve the topology of the image, through to stronger transformations that do not preserve topology, including random erasing, mix-up, and cut-mix. There are also photometric augmentations that act on the brightness values of each pixel to make the model invariant to color and contrast changes, forcing it to rely more on shape clues rather than spectral information (Tuli et al. 2021).

Dropout (Srivastava et al. 2014) works by temporarily disabling some neurons in a network layer during training, determined by a rate that specifies the probability of an individual neuron being deactivated. This approach reduces certain pathways' dominance and prevents co-adaptation among neurons, acting as a regularizer that decreases overfitting and simplifies the network structure, and particularly found useful in deep neural networks (DNNs) consisting of dense layers with numerous parameters or smaller training datasets (Baldi and Sadowski, 2014). However, it may not be ideal for convolutional neural networks (CNNs), in which maintaining the spatial structure of the input is crucial, and applying standard dropout can disrupt spatial coherence due to its random deactivation of individual neurons. Spatial Dropout (Tompson et al. 2015) addresses this by deactivating entire feature maps from the output of the previous layer, thus preserving the spatial coherence of the network's activations. As a regularizer, dropout is applied exclusively during training, with all neurons being active during inference (or the model's evaluation phase). The dropout concept inspired several adaptations for different purposes.

Monte Carlo dropout (Gal and Ghahramani, 2016), a Bayesian method for variational inference, employs dropout layers not only during training, but also during inference, to approximate the uncertainty in model predictions (Shridhar et al., 2019). By using dropout at inference time, multiple predictions are made with varying network configurations, leading to a distribution of outputs. These outputs are then aggregated (by averaging or majority voting) to provide a robust prediction and an uncertainty measure, typically the class-wise standard deviation of these predictions (Kendall et al. 2015). Kendall and Gal (2017) distinguish between two main types of uncertainty: aleatoric, inherent in data due to noise, and epistemic, stemming from limitations in a model's learned knowledge due to limited training samples. Dechesne et al. (2021) apply these concepts in a practical setting by developing a compound metric that merges the entropy of prediction distribution with the mutual information between the prediction and posterior over network weights. This metric effectively assesses both aleatoric and epistemic uncertainties, which the authors used along with the prediction-reference agreement to create qualification maps for analyzing network decisions in tasks such as extracting building footprints from benchmark datasets. MC-dropout has been used to quantify the uncertainty of deep models (Mukhoti and Gal, 2018) and to increase prediction robustness by improving model repeatability (Peng et al. 2021).

In this study, we introduce a novel workflow that leverages input normalization, Monte Carlo Dropout (MC-dropout), and careful tuning of the chosen loss function to enhance the temporal generalization capabilities of field boundary masks at a national scale. This approach significantly reduces the need for multi-year sample collection, pointing the way toward a cost-effective solution for large-scale, annually repeatable agricultural monitoring. Our focus is on producing cropland masks for annual crops, excluding perennial herbaceous and woody crops, aligning with common practices in the literature (Waldner et al. 2016, Potapov et al. 2022). The cropland masks are particularly designed to distinguish between the field interior, field edge, and non-field background classes, which improves the ability to perform post-hoc instance segmentation using the score maps for the field interior class.

## 2. Data and Study Area

The focal region of our study is Ghana (240,000 km²), which has a diverse agricultural landscape ranging from primarily rain-fed cereal cropping in the northern savanna regions to tree crop-dominated areas in the humid forests of the southwest (Sova et al. 2017). Agricultural fields in Ghana are typically small, averaging less than 2 hectares in size, and characterized by heterogeneous and often indistinct field patterns (Persello et al. 2019, Estes et al., 2022). Moreover, shifting agriculture is a common practice in this region (Kansanga et al. 2019). These agronomic factors, along with the frequent cloud cover, pose significant challenges in producing multi-year cropland maps of Ghana.

For the creation of annual cropland maps spanning the years 2018-2022, we used high-resolution (3.7 - 4.8 m) imagery derived from daily PlanetScore imagery. These images have four bands spanning the visible and NIR spectra and were compiled using two distinct methodologies. Initially, for the year 2018, a weighted temporal averaging approach was adopted to integrate daily imagery from November or December 2018 through February 2019 into a dry season temporal composite, as detailed in Estes et al. (2022). These composites were structured within tiles of 2000×2000 pixels (0.05° × 0.05° degree, n=8,116), each with an approximate resolution of 3 m (0.000025°). For the subsequent years of 2019 to 2022, we used Planet analytic base map imagery provided by Norway's Climate and Forests Initiative (NICFI[1]) at a resolution of 4.77 m, which is made from the best image during the time period (typically one month for 2020 onwards, and 6 months for earlier years), based on the cloud coverage and image quality using Planet's "best-on-top" algorithm. The collection of base map mosaics covered periods from June to December 2019, and November for the years 2020, 2021, and 2022. To ensure consistency across all years, these base maps were resampled to match the tiling grid and resolution used for the 2018 imagery. Additionally, to minimize boundary effects during the prediction phase, tiles were reprocessed to overlap, such that input dimensions are 2358×2358 pixels, with final predictions cropped to the original non-overlapping 2000x2000.

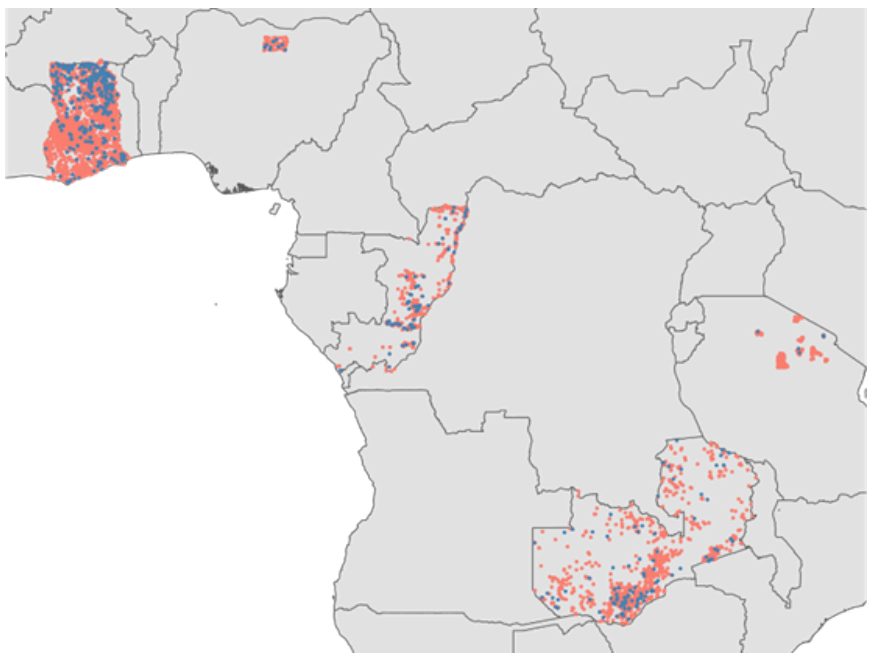

Figure 1. Spatial distribution of training (red) and validation datasets (blue) sampled in South-southwest Africa.

To train our cropland mapping model, we assembled a set of 4977 labeled images developed through manual digitization of field boundaries in the 2018 imagery, primarily as part of a prior mapping initiative (Estes et al., 2022). This dataset includes 4,229 labeled samples encompassing four different areas across the Ghanaian landscape where annual crops such as maize are primarily produced and further enriched the dataset with an additional 100 samples from Nigeria, 70 samples from Congo, and 578 samples from Tanzania derived from a similar procedure in 2020 to broaden the range of agronomic diversity, which was split into 4,781 samples for training, and the remaining 4% reserved for model validation (Figure 1).

---

[1] NICFI. URL https://www.nicfi.no/

Label polygons were converted into 200×200 pixel masks to denote the field interior, field boundary, and non-field areas, aligned with the dimensions of a 0.005° labeling grid. The delineation of class boundaries from the labels was executed by creating buffers around the boundaries of the original geometries with a thickness of 2 pixels.

To align with the corresponding PlanetScope imagery chips of 224×224 pixels, another buffering procedure with constant values was implemented, extending the dimensions of the labels. This adjustment ensured a harmonized input to the model, accommodating the 32x downsampling factor of our encoder-decoder model but the buffered area is masked out of the loss calculation during the training. It is noteworthy that the labels are relatively sparse, with few sample chips having more than 50% field pixels and around 20% negative chips with no crop fields (Figure 2).

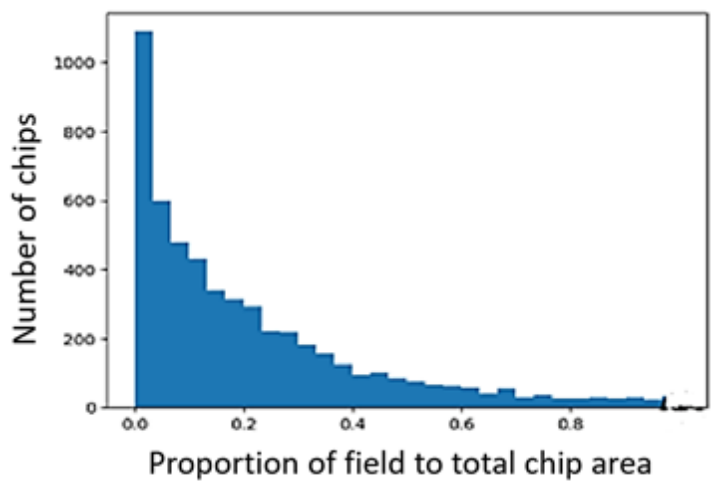

Figure 2. Distribution of the proportional field coverage in samples from the training dataset.

## 3. Methodology

We developed a methodology for national-scale mapping of agricultural fields using historical samples, integrating techniques to enhance temporal generalization, in order to eliminate the need for extensive annual data collection. Central to these techniques is the choice of input normalization and photometric augmentation, which were selected based on their ability to improve the model's generalization capabilities.

In our normalization process, we evaluated both min-max normalization and z-value standardization, applying these techniques across four distinct combinations to compute the necessary statistics, taking into account both the locality of the data and the spectral bands involved. Specifically, we calculated the statistics for each chip locally across all bands (local tile across all bands, *lab*); for the entire dataset across all bands (global across all bands, *gab*); and for each chip on a per-band basis (local tile per band, *lpb*); and for the entire dataset on a per-band basis (global per band, *gpb*). To further bolster the model's resilience against overfitting, and to enhance its adaptability to varying crop patterns and image reflectance artifacts, we expanded our training dataset with a combination of geometric and photometric transformations, thereby augmenting data diversity and robustness. The augmentations were applied on-the-fly with a 50% chance and chained with the order flip, rotation, resize, and photometric transformations. Flip was randomly selected from one of the horizontal, vertical, or diagonal types, and for photometric augmentation, one of the gamma correction, Gaussian noise, additive, and multiplicative noise were randomly applied in each iteration.

We adopted U-Net (Ronneberger et al., 2015) for our model, chosen for its simplicity, straightforward implementation, and reliability of predictions in land cover mapping (Anagnostis et al., 2021; Karra et al., 2021; Liu et al., 2022). Our U-Net variation employs a VGG-like architecture (Simonyan and Zisserman, 2014) with 12 convolutional layers and a 32x downsampling factor, producing feature outputs at each encoder stage of 64, 128, 256, 512, 1024, and 2048. This design strategically emphasizes the network's width over depth to increase the model's capacity, accommodating the limited label dimensions of 200×200 pixels, thus optimizing for our specific data constraints.

After experimenting with conventional and spatial dropout and different configurations of placing the dropout layer, we decided to use spatial dropout to regularize the model and added it to each convolution block in both the encoder and decoder subnetworks of the U-Net. We further applied MC-dropout (Gal and Ghahramani, 2016) to make model prediction ensembles, which besides providing an uncertainty measure, improved the model's generalization power and robustness, as shown in Table 2 in the results section. Through experimentation, we set the number of MC trials to 10 and used a fixed dropout rate of 0.15 for the training phase and 0.1 for the inference phase.

To refine the training of our model we adopted several strategies. We employed Focal Tversky loss (Abraham and Khan, 2019), known for its efficacy in handling imbalanced datasets and small object sizes. This is done by setting a weighting scheme ($\alpha$ and $\beta$ hyperparameters) that controls the trade-off between false positives (FP) and false negatives (FN) and the focal hyperparameter ($\gamma$), which controls the model's focus on hard-to-classify examples. We experimentally set $\alpha$ and $\gamma$ hyperparameters to 0.65 and 0.9 respectively, optimizing the model's ability to learn from challenging cases and reducing the impact of easy negatives. We implemented a dynamic class weighting scheme, based on an inverse frequentist approach, where weights are calculated on-the-fly for each class within a given input batch, as opposed to static weighting for the entire dataset. Furthermore, introducing object boundaries as a distinct class, provided a straightforward yet effective technique to enhance the model's ability to delineate individual fields. While these boundary delineations proved useful during training for field separation, they are excluded from the final predicted field mask, which has been previously shown to be effective (Garcia-Pedrero et al., 2019).

We developed our pipeline using the PyTorch 1.9.0 library and trained our large network (157 M parameters) on an A30 GPU machine for 120 epochs with a batch size of 32. After running initial experiments on SGD, SGD with momentum, Nesterov, Adam, and Sharpness-Aware Minimization (SAM) Optimizers (Foret et al. 2020) we adopted Nesterov as the optimizer in our pipeline. The initial learning rate was set to 0.003 which was updated with a polynomial learning rate decay policy with a power of 0.8.

After completing the training phase, we conduct a multi-faceted evaluation of the model using a selected set of metrics on the evaluation dataset. This includes precision, reflecting the model's accuracy in identifying field pixels; recall, measuring the ability to capture all actual field pixels; intersection-over-union (IOU), assessing the overlap between predicted and actual field areas for boundary accuracy; and F1-score, which harmonizes precision and recall, crucial for models dealing with imbalanced classes.

## 4. Results and Discussion

We produced annual maps of Ghana's croplands for 2018-2022 from the training samples that were predominantly collected from 2018 imagery, without fine-tuning on samples collected in subsequent years. To quantify the evaluation process, we randomly selected 4 tiles of size 2358x2358 from the 5 years and manually annotated the crop fields in the 20 scenes from the same imagery. Examination of spectral band distributions across this dry season period is shown in Figure 3, revealing a consistent mean within the visible (RGB) spectrum for the years 2019-2022, suggesting a relative constancy in the configuration of the major land cover and uses, soil exposure levels, and additional surface attributes contributing to mean spectral reflectance. The variability in the Near-Infrared (NIR) spectrum is likely attributable to inter-annual fluctuations in soil moisture content. The notable divergence in the brightness value distribution for the year 2018 is ascribed to variations in the data acquisition timing and the difference in the initial preprocessing method employed. The fluctuations in standard deviation highlight the underlying variability and complexity of the landscape, which could be influenced by any combination of the drivers for temporal shift (e.g. seasonal effects, soil moisture and atmospheric conditions, and the residual effects of vegetation dynamics) and intrinsic instability of radiometric calibration in the PlanetScope product (Frazier and Hemingway, 2021).

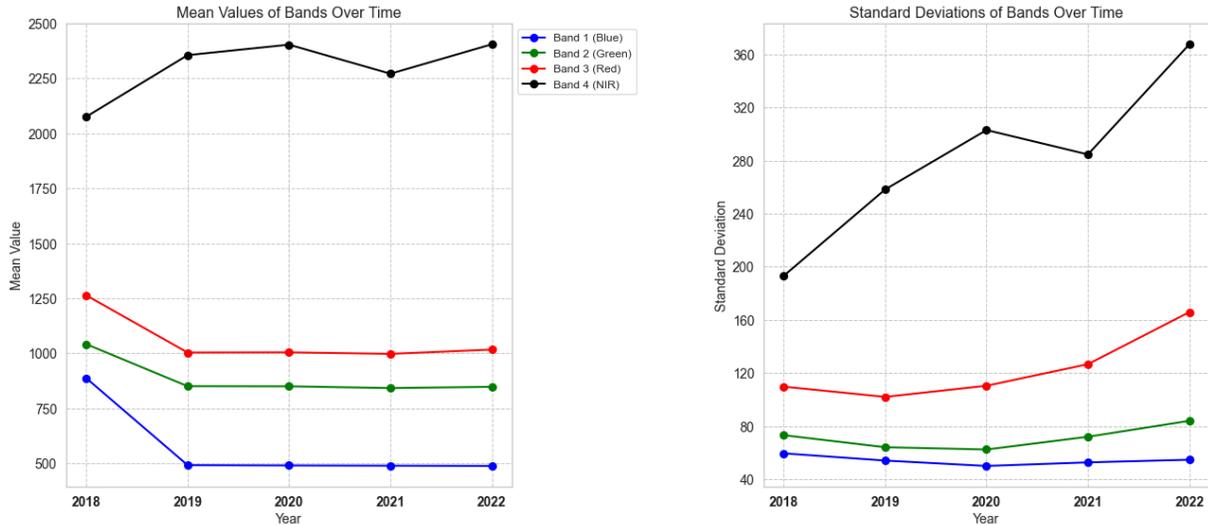

Figure 3. The means and standard deviation of image brightness values in each band for each year before normalization in the test dataset.

In our experiments to mitigate this domain shift, we found the choice of input normalization substantially influenced the generalization capacity of the model. Comparative analyses revealed that normalization techniques using min-max scaling with "lab" and "gab" conventions yielded the best and second-best outcomes, respectively for the tested dataset, challenging the prevailing consensus within the extant literature that advocates for band-specific (gpb or lpb) normalization (Syrris et al., 2019, Jeon et al., 2021, Carpentier, 2021, Khan et al. 2023). Nonetheless, our findings indicate that no singular normalization method consistently performed best across varying temporal and geographical contexts (Table 1). Our evaluation also showed that while all eight normalization techniques maintained the inter-band relational integrity (similar pairwise Pearson correlation coefficient), the "lab" and "gab" methods additionally preserved the original brightness value distributions of the imagery, whereas "lpb" and "gpb" approaches were prone to generate more pronounced extremes in the data values (Figure 4).

Table 1. Results of accuracy assessment on the field class using a small validation set consisting of 4 tiles of size 2358x2358 pixels from 2018 to 2022 (20 tiles in total) using the model trained on samples from 2018. All hyper-parameters are fixed except the input normalization strategy. The best results are shown in bold and the second best results are underlined.

a) Accuracy metric of different normalization procedures over the multi-temporal test dataset

| Normalization type | Precision | Recall | F1-score | IoU |
|---|---|---|---|---|
| mm-lab | 74.94% | 50.01% | **59.99%** | **42.84%** |
| mm-lpb | 63.89% | 52.22% | 57.47% | 40.32% |
| mm-gab | 71.60% | 51.53% | <u>59.93%</u> | <u>42.79%</u> |
| mm-gpb | <u>83.01%</u> | 41.32% | 55.18% | 38.10% |
| zv-lab | **85.42%** | 38.84% | 53.40% | 36.42% |
| zv-lpb | 77.33% | 35.73% | 48.87% | 32.34% |
| zv-gab | 50.78% | **56.45%** | 53.47% | 36.49% |
| zv-gpb | 59.21% | <u>53.96%</u> | 56.46% | 39.33% |

b) Accuracy metric of different normalization procedures over the multi-temporal test dataset separated by year

| Normalization type | | 2018 | 2019 | 2020 | 2021 | 2022 |
|---|---|---|---|---|---|---|
| mm-lab | IoU | **51.77%** | 42.76% | <u>34.02%</u> | <u>41.44%</u> | **44.15%** |
| | F1 | **68.22%** | 59.89% | <u>50.77%</u> | <u>58.57%</u> | **61.26%** |
| mm-lpb | IoU | 49.65% | **48.49%** | 32.06% | 37.81% | 36.16% |
| | F1 | 66.36% | **65.31%** | 48.55% | 54.87% | 53.12% |
| mm-gab | IoU | 50.32% | <u>44.08%</u> | **37.69%** | **44.33%** | <u>38.82%</u> |
| | F1 | 66.95% | <u>61.19%</u> | **54.75%** | **61.43%** | <u>55.93%</u> |
| mm-gpb | IoU | <u>50.58%</u> | 34.15% | 29.74% | 39.29% | 36.24% |
| | F1 | <u>67.18%</u> | 50.91% | 45.84% | 56.41% | 53.20% |
| zv-lab | IoU | 48.99% | 36.09% | 22.38% | 36.45% | 36.68% |
| | F1 | 65.76% | 53.04% | 36.58% | 53.43% | 53.67% |

| | | | | | |
|---|---|---|---|---|---|
| zv-lpb | IoU | 46.98% | 35.39% | 17.38% | 30.67% | 31.18% |
| | F1 | 63.93% | 52.28% | 29.61% | 46.95% | 47.54% |
| zv-gpb | IoU | 48.44% | 42.57% | 31.20% | 41.15% | 36.56% |
| | F1 | 65.26% | 59.72% | 47.56% | 58.31% | 53.54% |
| zv-gab | IoU | 45.75% | 36.68% | 31.54% | 35.91% | 35.39% |
| | F1 | 62.78% | 53.67% | 47.96% | 52.85% | 52.27% |

c) Accuracy metric of different normalization procedures over the multi-temporal test dataset separated by geography

| Normalization type | | Tile 1 (id: 487103) | Tile 2 (id: 513911) | Tile 3 (id: 513254) | Tile 4 (id: 539416) |
|---|---|---|---|---|---|
| mm-lab | IoU | **45.70%** | <u>46.56%</u> | **43.31%** | 36.37% |
| | F1 | **62.73%** | <u>63.53%</u> | **60.44%** | 53.34% |
| mm-lpb | IoU | 29.52% | **47.61%** | 37.54% | 37.04% |
| | F1 | 45.59% | **64.51%** | 54.59% | 54.05% |
| mm-gab | IoU | 39.92% | 46.47% | <u>39.58%</u> | **46.47%** |
| | F1 | 57.06% | 63.45% | <u>56.71%</u> | **63.45%** |
| mm-gpb | IoU | 40.44% | 38.86% | 36.73% | 36.71% |
| | F1 | 57.59% | 55.97% | 53.73% | 53.70% |
| zv-lab | IoU | <u>41.80%</u> | 35.36% | 35.28% | 35.36% |
| | F1 | <u>58.96%</u> | 52.25% | 52.16% | 52.25% |
| zv-lpb | IoU | 37.29% | 28.10% | 34.12% | 28.10%  |
|        | F1  | 54.32% | 43.88% | 50.88% | 43.88%  |
| zv-gpb | IoU | 29.41% | 45.68% | 35.28% | <u>37.45%</u> |
|        | F1  | 45.45% | 62.71% | 52.15% | <u>54.49%</u> |
| zv-gab | IoU | 27.08% | 46.40% | 36.90% | 28.23%  |
|        | F1  | 42.62% | 63.39% | 53.90% | 44.02%  |

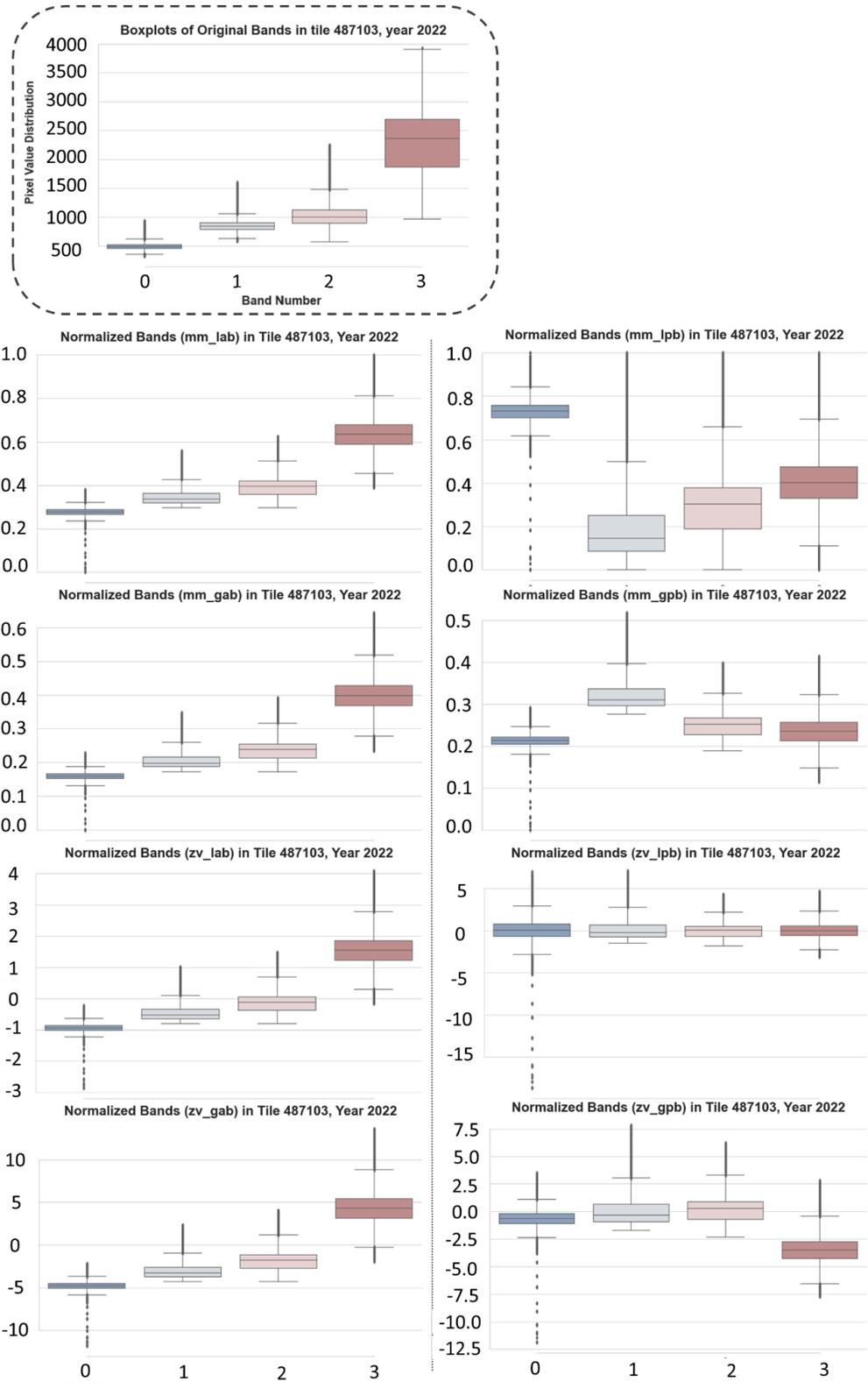

Figure 4. Distribution of the original bands and different normalization procedures for a single image tile in the year 2022.

Through experimentation, we observed that a higher capacity U-Net demonstrates improved performance, and notably, increasing the model's width proved to be more beneficial for enhancing capacity than deepening its layers. The non-regularized model with no dropout or photometric augmentation, exhibited superior performance in generating crop maps for the year 2018. However, it was unable to accurately predict cropland distributions in subsequent years, characterized by significant rates of omission errors across various regions. Incorporating photometric augmentation into the model's training regimen improved true positive (TP) rates but also increased false positives (FP). Further integration of the dropout layers enhanced the model's generalization capabilities, effectively reducing the incidence of FP pixels, but in the absence of photometric augmentation, using dropout markedly elevated the false negative (FN) rate across all evaluated years. Notably, the application of MC-dropout emerged as the most effective strategy, as evidenced by the empirical results presented in Table 2 and Figure 5 (and Appendix), indicating a marked improvement in model performance. Optimal results were achieved with a 0.15 training dropout rate, and a 0.1 prediction dropout rate, with the number of MC trials set to 30. We also tried matching the histogram of the subsequent years (2019-2022) with the training year (2018) as a test phase augmentation with our best model, but this approach did not improve outcomes (table 2.3). We found TF loss to perform superior on all the tested metrics compared to CE loss with a large margin and calculating the class weights locally produced much lower FN compared to global class weights for weighted loss calculations. However, lowering the capacity of the model to half (e.g. 80 M trainable parameters) only slightly increased the number of FN pixels and can be used if the computation resources are limited (table 3).

Table 2. Comparison of the effects of different combinations of photometric augmentation and dropout on mitigating the temporal domain shift. The values in parentheses represent the metrics before adaptive thresholding when probability scores are hardened with the fixed value of 75.

| Experiment | 2018 | 2019 | 2020 | 2021 | 2022 | Across all years |
|---|---|---|---|---|---|---|
| No MC-dropout, No photo aug | IoU: 51.71% F1: 68.17% | IoU: 23.83% F1: 38.49% | IoU: 11.73% F1: 21.01% | IoU: 17.70% F1: 30.08% | IoU: 23.42% F1: 37.96% | IoU: 25.41% F1: 40.52% |
| MC-dropout, No photo aug | IoU: _53.38_% (44.71) F1: _69.60_% (61.79) | IoU: 39.01% (18.23) F1: 56.12% (30.84) | IoU: 24.78% (8.12) F1: 39.72% (15.03) | IoU: 33.07% (14.96) F1: 49.70% (26.03) | IoU: 41.09% (18.82) F1: 58.24% (31.68) | IoU: 38.45% (20.75) F1: 55.54% (34.37) |
| No MC-dropout, photo aug | IoU: **54.39**% F1: | IoU: 32.23% F1: | IoU: 28.11% F1: | IoU: 31.13% F1: | IoU: 33.29% F1: | IoU: 34.03% F1: |
|  | **70.46**% | 48.75% | 43.88% | 47.48% | 49.96% | 50.79% |
| Only train dropout, photo aug | IoU: 51.80% F1: 68.25% | IoU: 41.52% F1: 58.68% | IoU: **35.45**% F1: **52.35**% | IoU: 39.61% F1: 56.74% | IoU: 43.87% F1: 60.98% | IoU: 42.33% F1: 59.48% |
| Both MC-dropout, photo aug | IoU: 51.77% (50.26) F1: 68.22% (66.89) | IoU: **42.76**% (36.08) F1: **59.89**% (53.02) | IoU: 34.02% (25.70) F1: 50.77% (40.90) | IoU: **41.41**% (32.72) F1: **58.57**% (49.31) | IoU: **44.15**% (40.0) F1: **61.26**% (57.1) | IoU: **42.84**% (36.97) F1: **59.99**% (53.98) |

Table 3. Comparison of the effects of model capacity, loss function, and histogram matching against the best model with all augmentations, TF loss, and MC-dropout. All hyper-parameters except the property under study are the same between experiments.

| Experiments | Precision | Recall | F1-score | IoU |
|---|---|---|---|---|
| Best model | 74.94% | 50.01% | 59.99% | 42.84% |
| Half-capacity (width) | 75.73% | 48.88% | 59.41% | 42.26% |
| TFL-Global weight | 81.80% | 32.49% | 46.51% | 30.30% |
| CE-local weight | 67.88% | 26.33% | 37.94% | 23.41% |
| Histogram matching | 58.47% | 55.13% | 56.75% | 39.62% |

We noticed the most substantial distinction was in the probability maps generated by the two dropout approaches, with the conventional method exhibiting an overconfidence in its predictions and has a much smaller range for the probability of the positive class. However, if the proper hardening threshold is used, spatial dropout is more efficient in reducing false positives (FP) relative to traditional dropout layers. Furthermore, our analysis revealed that the normalization procedure also significantly influences the model's prediction confidence. Specifically, prediction scores generated from the mm-lab normalization and spatial MC-dropout exhibited the widest range in probability values, making it an invaluable asset for enhancing the training dataset. This feature is in addition to the utility of other layers, such as variation and mutual information, derived from Monte Carlo trials.

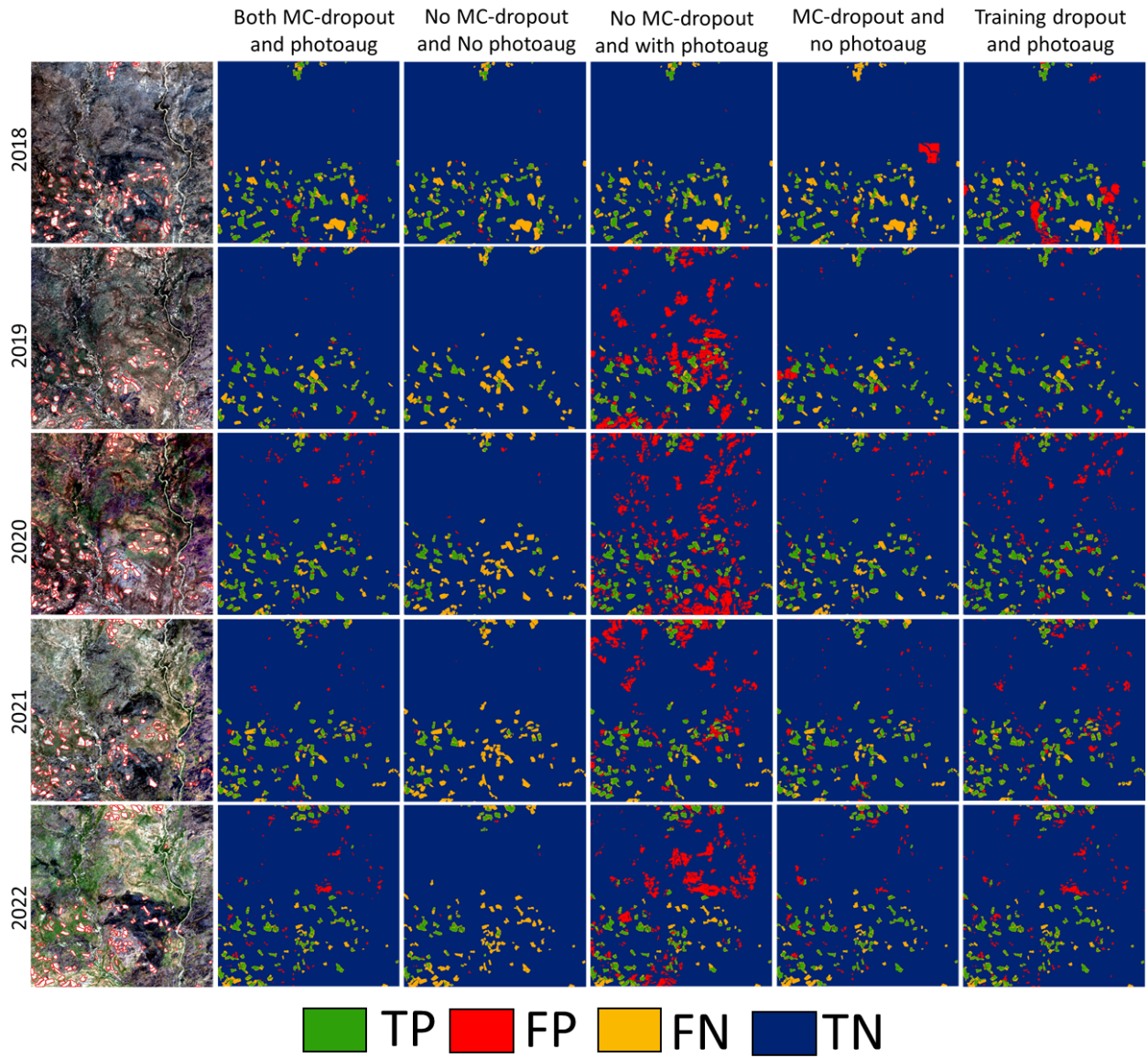

Figure 5. Spatial confusion matrix showing the effect of different combinations of spatial dropout and photometric augmentation for a sample prediction tile (487103) through the years 2018 to 2022.

## 5. Conclusion

This work demonstrates that careful choice of pre-processing (e.g. input normalization, and image augmentations), and tuning the capacity of the model accompanied by dropout regularization in both training and prediction phases significantly improves the generalization power of the model and its capability for temporal domain adaptation. This capability enabled a model trained primarily on samples for a single year, with imagery having a different provenance, to make high-resolution, multi-year maps of field boundaries in smallholder-dominated croplands at national scales, an important requirement for agricultural monitoring. Although the resulting maps still have some notable omission errors in each year, these errors were substantially reduced by the techniques used here, and more closely captured the inter-annual distribution of crop fields. The remaining error may be further reduced by fine-tuning with a small number of labels collected for each year, targeted using the uncertainty information provided by the MC trials, with the possibility of auto-generating labels from regions with low prediction uncertainty.

## Acknowledgements

Support for this research was provided by the Enabling Crop Analytics at Scale (ECAAS) project, funded by the Bill and Melinda Gates Foundation, with additional support from the National Science Foundation (Award #1924309) and Omidyar Network's Property Rights Initiative, now PLACE.

## Appendix

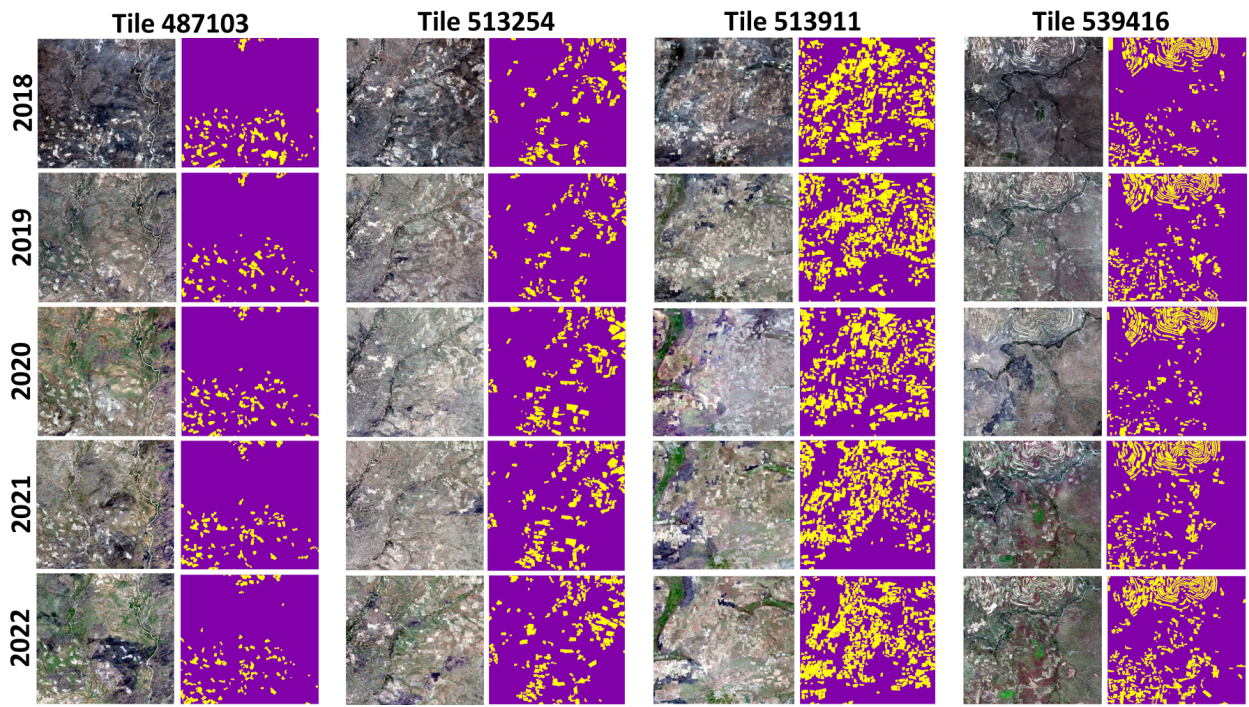

Figure A1. Overview of the prediction dataset used to evaluate the performance of the model in reusing historical data. It consists of 4 tiles of 2358x2358 across the span of years from 2018 to 2022 from different regions in Ghana.

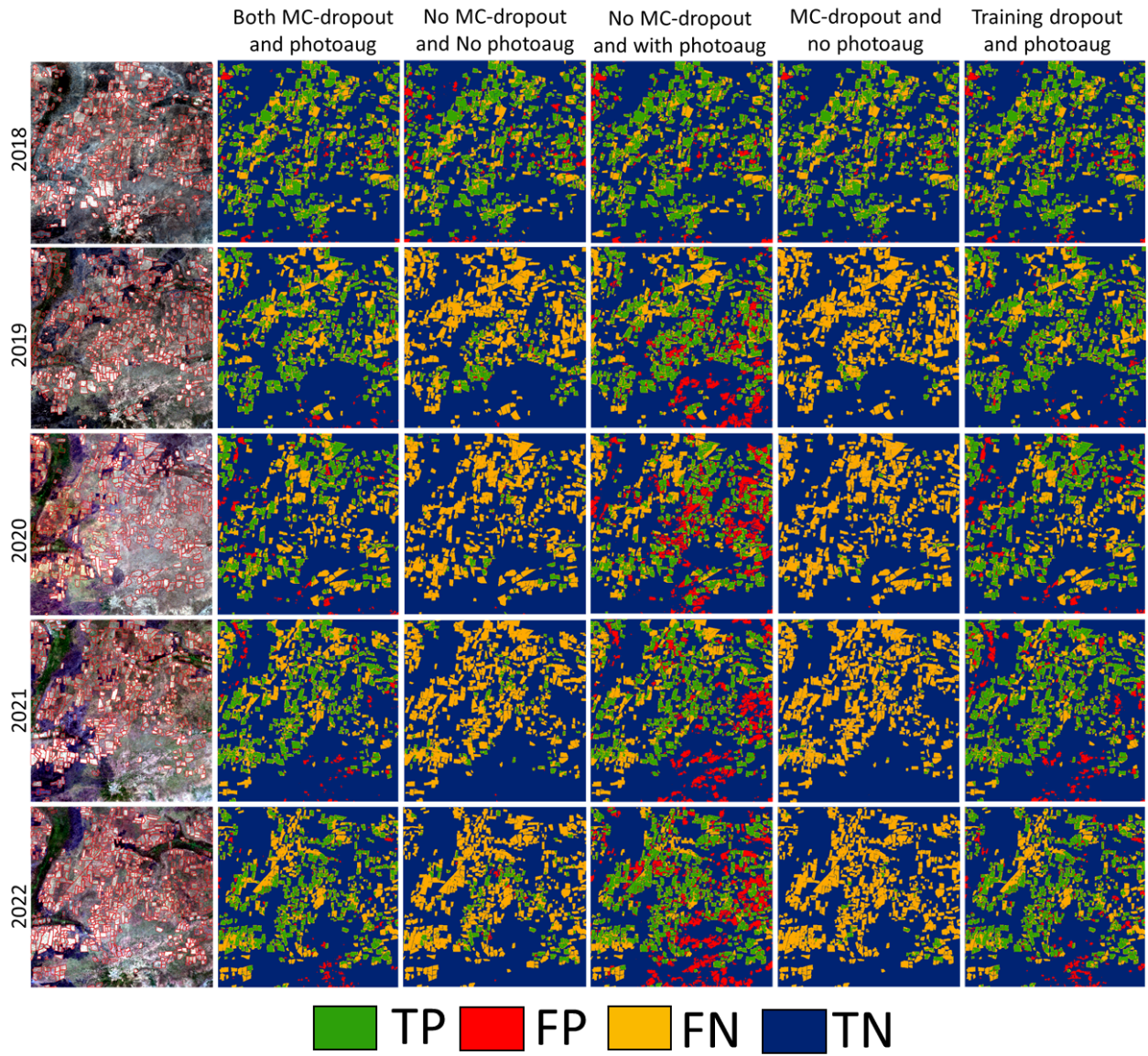

Figure A2. Spatial confusion matrix showing the effect of different combinations of dropout and photometric augmentation for tile 513911 through the years 2018 to 2022.

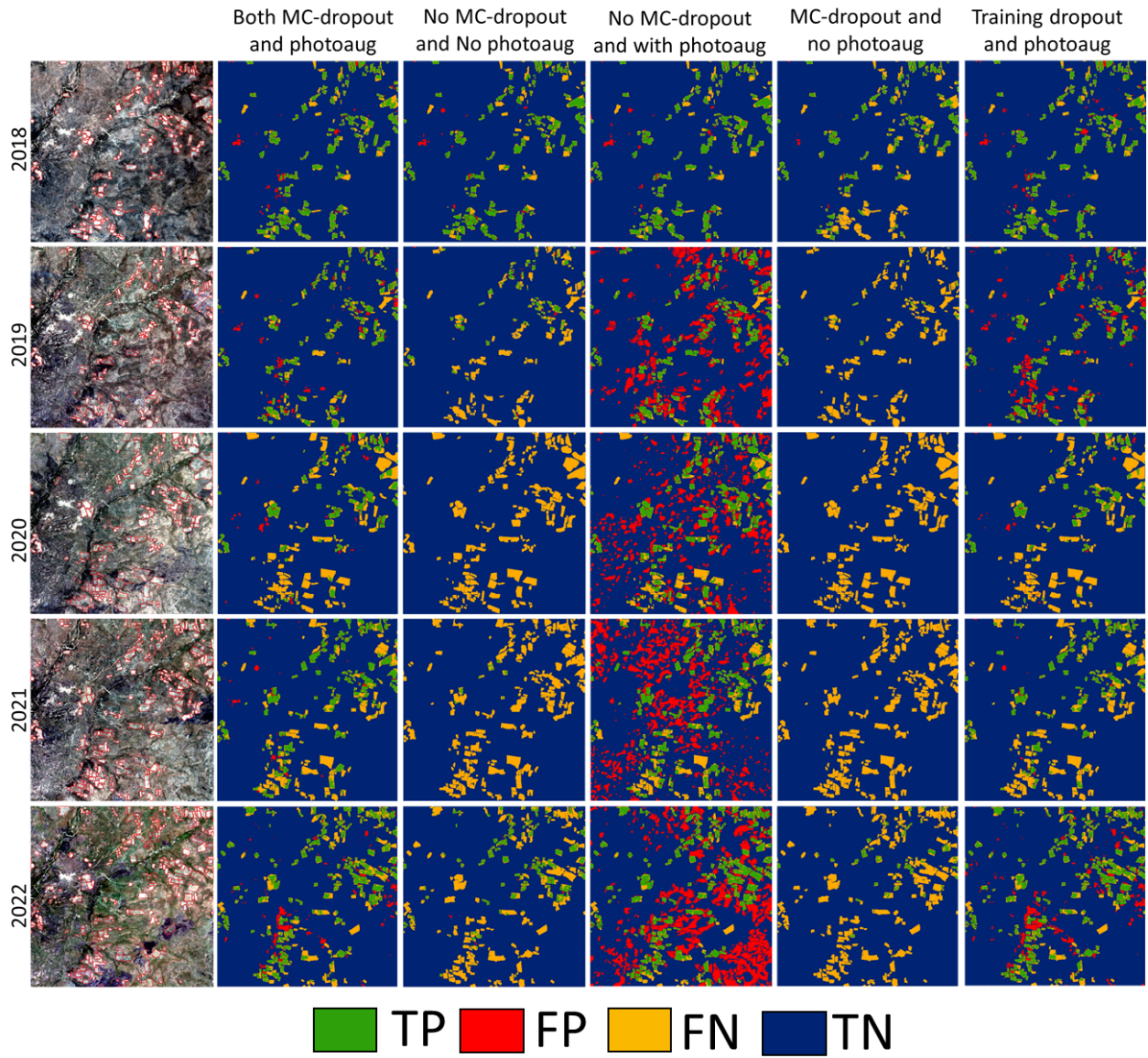

Figure A3. Spatial confusion matrix showing the effect of different combinations of dropout and photometric augmentation for tile 513254 through the years 2018 to 2022.

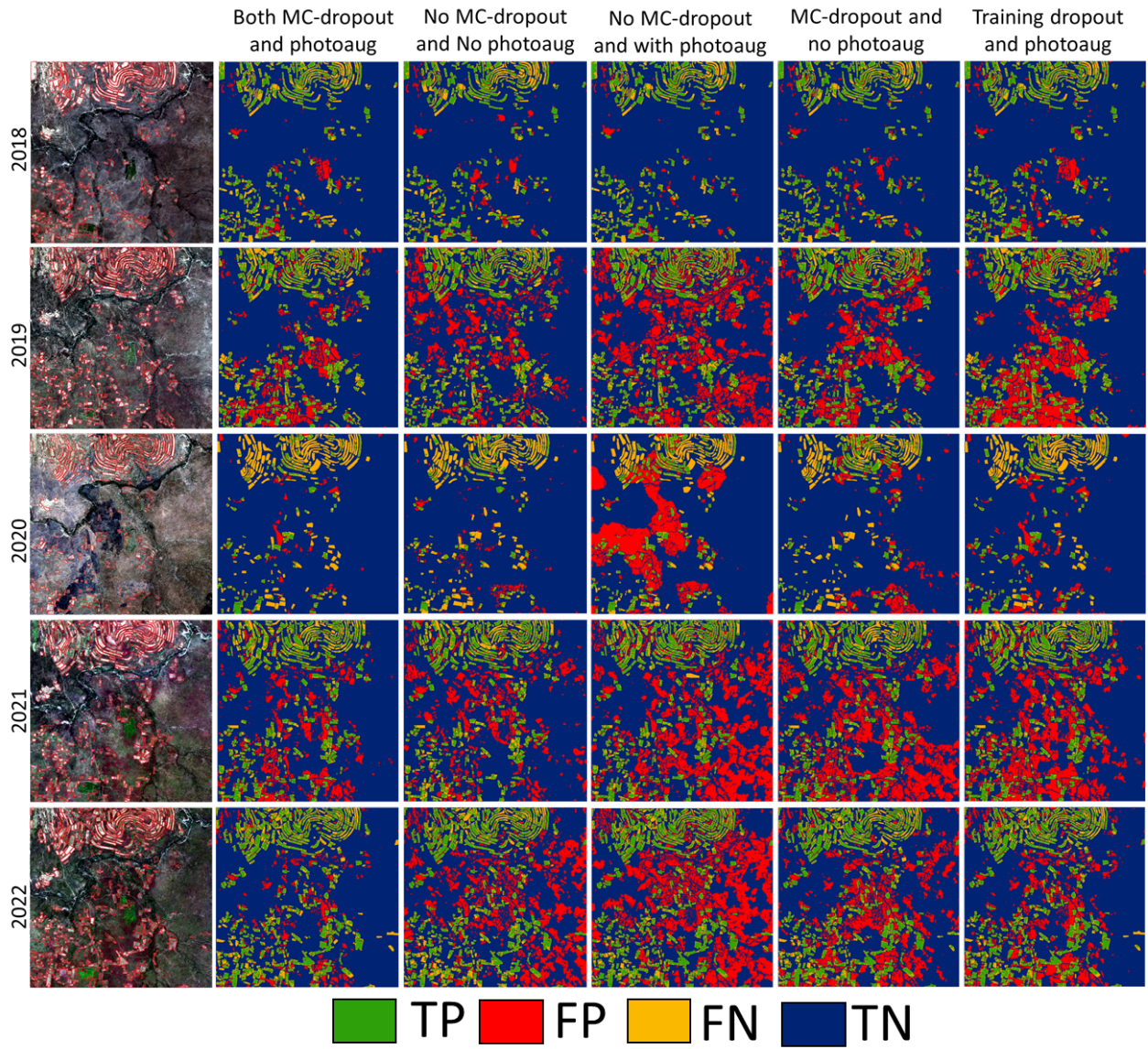

Figure A4. Spatial confusion matrix showing the effect of different combinations of dropout and photometric augmentation for tile 539416 through the years 2018 to 2022.